\documentclass[journal]{IEEEtran}
\usepackage[utf8]{inputenc}

\title{Intelligent Icing Detection Model of Wind Turbine Blades Based on SCADA data}
\author{Wenqian Jiang, Junyang Jin
\thanks{Wenqian Jiang is with China-EU Institute for Clean and Renewable Energy, Huazhong University of Science and Technology, Wuhan, China, 430074.}
\thanks{Junyang Jin is with HUST-Wuxi Research Institute, Wuxi, China, 214000.}
}

\usepackage{url}
\usepackage{booktabs}
\usepackage{amsfonts}
\usepackage{graphicx}
{}
{}
{}
\usepackage{algorithm}
\usepackage{algorithmicx}
\usepackage{algpseudocode}
\usepackage{threeparttable}
\usepackage{nomencl}
\usepackage{url}
\usepackage{amsmath} 
\usepackage{enumerate}
\usepackage{threeparttable}
\usepackage{multirow}
\usepackage{color}

\begin{document}

\maketitle

\begin{abstract}
Diagnosis of ice accretion on wind turbine blades is all the time a hard nut to crack in condition monitoring of wind farms. Existing methods focus on mechanism analysis of icing process, deviation degree analysis of feature engineering. However, there have not been deep researches of neural networks applied in this field at present. Supervisory control and data acquisition (SCADA) makes it possible to train networks through continuously providing not only operation parameters and performance parameters of wind turbines but also environmental parameters and operation modes. This paper explores the possibility that using convolutional neural networks (CNNs), generative adversarial networks (GANs) and domain adaption learning to establish intelligent icing diagnosis frameworks under different training scenarios. Specifically, parallel GANs framework combined with CNN (PGANC) and parallel GANs framework combined with transfer learning (PGANT) are proposed for sufficient and insufficient target wind turbine labeled data, respectively. The basic idea is that we consider a two-stage training with parallel GANs, which are aimed at capturing intrinsic features for normal and icing samples, followed by classification CNN or domain adaption module in various training cases. Model validation on three wind turbine SCADA data shows that two-stage training can effectively improve the model performance. Besides, if there is no sufficient labeled data for a target turbine, which is an extremely common phenomenon in wind farms, the addition of domain adaption learning makes the trained model show better performance. Overall, our proposed intelligent diagnosis frameworks can achieve more accurate detection on the same wind turbine and more generalized capability on a new wind turbine, compared with other machine learning models and conventional CNN-based methods.   

\end{abstract}

\section{Introduction}

Global warming is one of the most powerful actors for promoting renewable sources a promising alternative to traditional fossil fuels, in which wind energy has received increasing attention. As “the fastest growing renewable energy source in the world” with an average annual growth of 30\% over the past two decades \cite{habibi2019reliability}, wind power has grown to an essential part to achieve emission reduction targets in the future. The Global Wind Energy Council makes a prediction that the installed capacity global wind power is expected to increase to 840GW by the end of 2022 \cite{GWEC}.

In order to be competitive in the electrical energy market, renewable energy requires the advances in the reliability, availability, maintainability and safety  \cite{zhang2012international}, wind power is no exception. Wind power generation often faces some challenges that restrict its development, such as blades icing monitoring studied in our paper. Wind farms are mostly located on high altitude mountainous areas, making their operating conditions under low temperature and high humidity. Therefore, it’s almost unavoidable for wind turbine blades to be frozen, especially in winter. Once there is ice appearing on wind turbine blades, it would change the aerodynamic performance of the blades, yield into power generation loss, threaten the normal operation of wind turbines, and even lead to security risks \cite{gantasala2016influence}. Besides, the icing condition will be worse and worse, unless human intervention.

Various signals acquired by the monitoring technology and sensors can be utilized to distinguish the icing turbines. For example, a monitoring camera can be installed on the wind turbine \cite{homola2006ice}, and then researchers apply hyperspectral imaging technology to find icing wind turbine blades \cite{rizk2020hyperspectral}. Vibration signal analysis such as using changes in natural frequencies can contribute to detect damage for blades\cite{yu2018crack}. Besides, the Supervisory Control and Data Acquisition (SCADA) is a strong system for data acquisition and equipment monitoring, which is also
the most widely used and technologically advanced in fault diagnosis of a large amount of wind power elements\cite{tautz2016using, dai2018ageing, dao2018condition}. From SCADA, not only the operating parameters of wind power equipment, but also environmental parameters, such as temperature,wind speed, wind direction, are collected, both of which give a full description for the status of wind turbines. More and more researchers use it for data modeling and analysis to mine information of fault equipment and blade icing detection. \cite{dao2018condition, cao2016study}. 

Physical model of icing wind turbine blades and theoretical analysis is mainly considered in the existing methods. Besides, the International Electro technical Commission (IEC) has developed many standards and guidelines to help analyze faults in wind turbines\cite{IEC}. Battisti \textit{et al.} give a detailed description about the causes and types of blade icing and analyze the effect of icing process on wind turbines in \cite{battisti2015wind}. Lichun \textit{et al.} study the relationship between the icing positions on wind turbines and output power in \cite{shu2016influences}. Han \textit{et al.} focus on the freezing condition under different temperature, liquid water content and wind speed in \cite{han2012scaled}. Dong \textit{et al.} utilize the deviation degree of output power performance, mechanical performance and aerodynamic performance characteristic parameters in icing blades to establish detection model based on SCADA data \cite{dong2020blades}. To identify the power loss caused by icing blades, three methods of creating a power threshold are studied by Davies \textit{et al.} in \cite{davis2016ice}. The relationship between leaf icing and the power characteristics of wind turbines under natural conditions is explored by Shu \textit{et al.} in \cite{shu2018study}.

Data-driven models have shown promising performance in exploring the relationship between the operation variables and the icing conditions automatically. Li \textit{et al.} propose a method for identification
of icing blades based on logistic regression \cite{ningbo2018ice}. Skrimpas \textit{et al.} combine vibration and power curve analysis with the K-Nearest Neighbours (KNN) method for icing turbines detection \cite{skrimpas2016detection}. Zhang \textit{et al.} propose a blade icing prediction model based on Random Forest Classifier \cite{zhang2018ice}. Yuan \textit{et al.} put forward a deep time series classification model called WaveletFCNN for wind turbine blade icing detection \cite{yuan2019waveletfcnn}. Song \textit{et al.} propose a wind turbine monitoring system based on a Bayesian data-driven approach \cite{song2018wind}. An icing blades detection model based on random forest using 29 SCADA features is proposed by Ge \textit{et al.} in \cite{ge2017prediction}. Zhou \textit{et al.} also constructed features from SCADA data based on expert knowledge, and SVM with particle swarm optimization is adopted to build the icing detection model of wind turbine blades \cite{guangfei2018ice}.

The model-based methods mainly depend on physical and mathematical model of systems or elements under theory hypothesis, which are often based on rich knowledge background and working experiences. Their solutions are definite and interpretable, and the detection process is mature and systematic. However, there is a big gap between theoretical analysis and practical application. Due to the existence of various noise and other uncertain and uncontrollable factors in real working conditions, it is almost impossible to fully meet the assumptions of theoretical analysis, which leads to the unavailability of model-based analysis in some cases. In recent years, icing detection methods based on data-driven analysis is more popular with the development of neural network. Data mining from the signal information directly analyzes the uncertain and uncontrollable factors in real working conditions, gets rid of the hypothesis constraints of the theoretical conditions, and is easier to obtain useful icing detection criteria.

For data-driven methods, especially for icing detection methods based on machine learning and deep learning, high-quality labeled data is often the key to acquiring high diagnosis accuracy. Although there will be a large amount of data generated in the monitoring systems of wind farms, manually labelling data is always a big challenge, so it is difficult to get enough high-quality labeled samples for training. In addition, there is always much more normal samples than icing samples. The imbalanced case will force the neural networks to sacrifice the accuracy of minority samples to obtain higher accuracy of the whole testing samples. However, the minority samples (e.g. icing samples in this paper) are exactly what we need to pay attention to. The above phenomenon leads to the problem of insufficient labeled samples and imbalanced problem.

Aimed at above problems, two intelligent icing detection frameworks, parallel GANs framework combined with CNN (PGANC) and parallel GANs framework combined with transfer learning (PGANT), are proposed to help identify icing blades in wind farms using SCADA data. We firstly establish PGANC: two parallel GANs followed by a CNN. Through a two-stage training, it can achieve good performance on a single turbine. But when we test the trained PGANC model on a different wind turbine, the performance is not so satisfactory. Specifically, the identification of normal wind turbines is easier than that of icing cases. This is true because different wind turbines often operate under various environments and load conditions together with different physical characteristics, which means that the icing conditions are different from turbine to turbine. In addition, labeling the icing data from normal data is always time-consuming and labor-intensive. Therefore, we improve PGANC to PGANT: two parallel GANs followed by a transfer learning framework to deal with the above problem. This paper makes the following contributions:

\begin{enumerate}[1)]
    \item Two intelligent icing diagnosis frameworks: PGANC and PGANT are proposed for various training scenarios. For the same wind turbine with sufficient labeled data, PGANC shows better performance on the imbalanced SCADA data, where the number of icing samples is much less than that of normal samples. For a new wind turbine without sufficient labeled data, PGANT adopts domain adaption module to transfer the trained invariant features of another historical wind turbine.
    
    \item A two-stage training process is taken into consideration. In both PGANC and PGANT, we first train GANs and CNNs or GANs and transfer framework in separate series way. After acquiring the optimal parameters, based on which the second holistically training can improve the performance further and reduce computing time.
    
    \item Extensive experiments have been conducted to validate the effectiveness of our proposed methods. In PGANC, three wind turbine SCADA data is used. The results show that PGANC outperforms conventional CNNs, and the two-stage training is effective. Performance on 3 different transfer scenarios proves that PGANT can achieve more generalize capability.   
\end{enumerate}

This paper is organized as follows. Section II reviews related works including CNN and GAN for fault diagnosis and transfer learning. Section III proposes our two intelligent icing diagnosis frameworks. Experiment results and comparison are given in Section IV. Finally, conclusion and future work are drawn in Section V.

\section{Related Works}

In recent years, CNNs have long been popular in deep learning~\cite{cnn11,cnn12,cnn13}. The initial CNN architecture was proposed by LeCun et al. in works \cite{lecun1989handwritten} and \cite{lecun1989backpropagation}, of which main features are local connections, shared weights, and local pooling \cite{saxe2011random}. They have been widely used in classification tasks. For example, a 1-D convolutional neural network-based approach \cite{chen2020robust} has been proposed by Chen et al. to diagnose both known and unknown faults in rotating machinery under added noise. Gao et al.~\cite{gao2020semi} developed a semi-supervised learning method based on CNN for steel surface defect recognition. \cite{liu2016dislocated} proposed a dislocated time series convolutional neural network called DTS-CNN to dislocate the 1D input mechanical signal into an output matrix.  

In addition to CNNs, GANs occupy an increasingly pivotal position in the deep learning community. They were initially introduced by Goodfellow \textit{$et\ al.$} as an unsupervised machine learning method~\cite{goodfellow2014generative}.The basic idea of GANs is that it generates prototypical samples through a generator with random data points that satisfy a certain
distribution (e:g: Gaussian distribution). GANs are well known by the great power of feature extraction and pattern recognition~\cite{fgan,fgan1,fgan2}. In addition, GANs present outstanding performance in dealing with imbalanced datasets~\cite{yin2015empirical}, which is especially demanding to handle the relatively minor icing cases in our paper. Various adversarial algorithms have been developed to construct GANs. GANs were applied to tackle imbalanced fault classification problems of machinery~\cite{gan13}.  For more details, interested readers may refer to~\cite{ganzoo} for a comprehensive discussion of GANs and their variants. 

Through transferring the knowledge from source domain to target domain under diverse operating conditions, transfer learning is often regarded as an efficient tool to solve the diagnostic or predictive tasks of unlabeled or insufficient data. We adopt the definition of transfer learning in \cite{cheng2019wasserstein}:

Transfer learning is proposed with the aim to learn a prediction function $\mathcal{F}(x)$ : $x \rightarrow y$ for a learning task $\mathcal{T}_t$ by leveraging knowledge from source domain $\mathcal{D}_s$ and $\mathcal{T}_s$, where $\mathcal{D}_s \neq \mathcal{D}_t$ or $\mathcal{T}_s \neq \mathcal{T}_t$. In most of the cases, $\mathcal{D}_s$ contains a much larger
dataset than $\mathcal{D}_t$.

How to measure the distribution similarity between source domain and target domain is of vital importance in transfer learning. MMD is the most common seen distance measure used in existing transfer networks \cite{long2015learning, gretton2012kernel, sejdinovic2013equivalence}.The MMD metric
is a special case of the integral probability metrics which measures the distance between two probability distributions via mapping the samples into a Reproducing Kernel Hilbert Space associated with a given kernel.

\section{Wind Turbine Blade Icing Detection using Deep Learning}

\begin{figure*}[h!]
\centering
\includegraphics[scale=0.5]{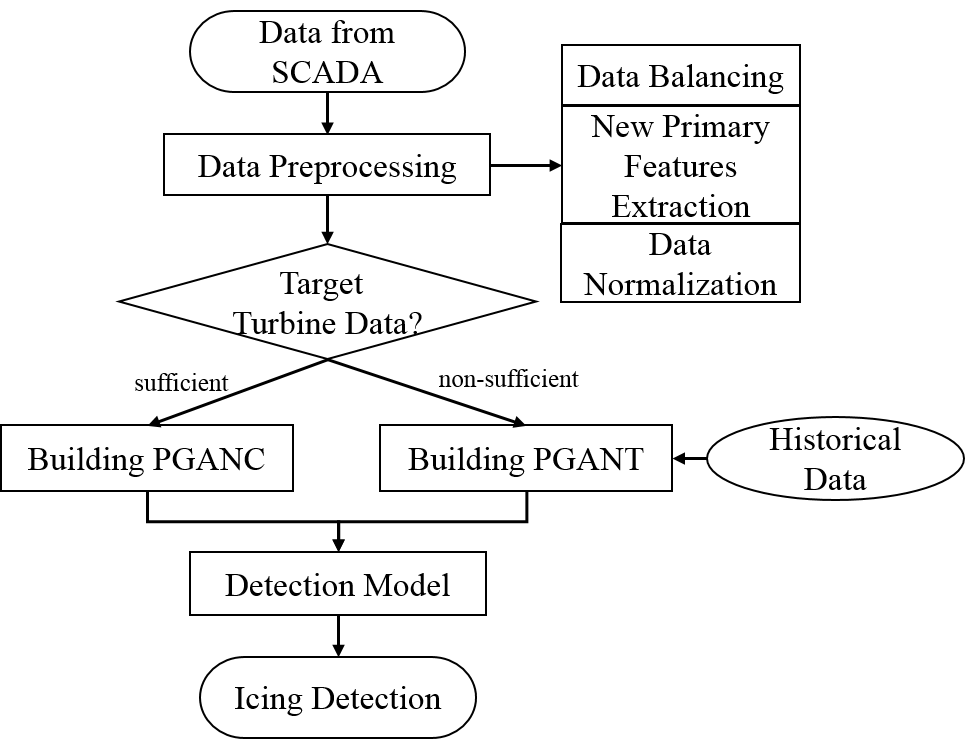}
\caption{Flow chart of model construction}
\label{fig:W_f}
\end{figure*}

\subsection{PGANC: sufficient labeled data scenario}
\label{tit:f1}
\subsubsection{Method overview}
For the same turbine with sufficient labeled data, the training process is supervised. Our neural network designed to classify states of wind turbine blades is as follows:
\begin{equation}
    \begin{aligned} 
    Y = NN(x)
    \end{aligned}
    \label{wn}
\end{equation}

where $x\in \mathcal{R}^{1\times28}$ is a vector containing 28 variables, $Y\in \mathcal{R}^{1\times2}$ is the output of the network indicating the probability of normal and icing state, and  $NN(\cdot)$ is the neural network described by a composition of nonlinear functions that maps inputs to outputs.

\begin{figure}[h!]
\centering
\includegraphics[scale=0.28]{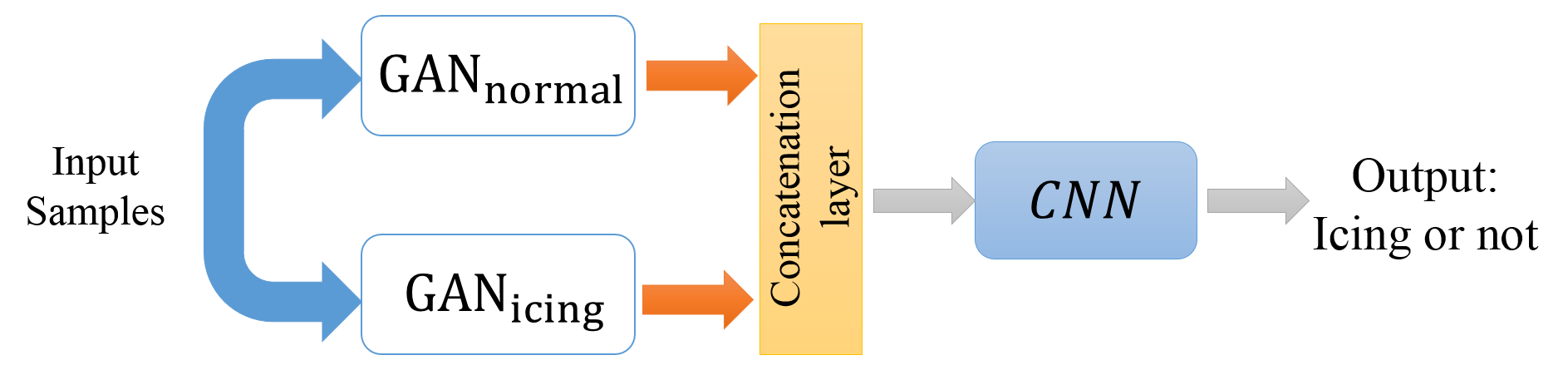}
\caption{GAN+CNN Framework}
\label{fig:gan_cnn_f}
\end{figure}

PGANC is composed of two GANs in parallel following by a CNN classifier as shown in Fig \ref{fig:gan_cnn_f}. As an unsupervised learnning technique, GANs are powerful in extracting hidden features from inputs and capturing inherent patterns for target classes. CNNs are capable of learning spatial information among different dimensions and across multiple channels of inputs~\cite{khan2019survey}. By combining GANs and CNNs, the derived neural network can efficiently conduct feature extraction and thus accurately perform classification. 

A GAN consists of a generator $G(\cdot)$ and a discriminator $D(\cdot)$. The generator transform input samples into synthetic ones in order to deceive the discriminator. The generator applies an encoder to extract low dimensional features from inputs which are used by the decoder as ingredients to craft a fake work. Somehow, the encoder and decoder are anagolous to recipes for decomposition and synthesis, respectively. Hence, a generator learns unique recipes for the target class if it is trained purely using the samples from that class. Provided a sample from a different class, the generator will produce a pool forgery because of the wrong recipes used. In contrast, the discriminator aims to determine whether a given sample is genuine. It applies an independent encoder to explore major features of samples as the basis for judgement. These features can be treated as signatures that mark genuine and synthetic samples. Similar to the generator, the discriminator cannot extract correct signatures unless input samples come from the target class. In this case, GANs are highly effective in recognizing patterns of different classes~\cite{jiang2019gan}.

\begin{figure}[ht!]
\centering
\includegraphics[scale=0.28]{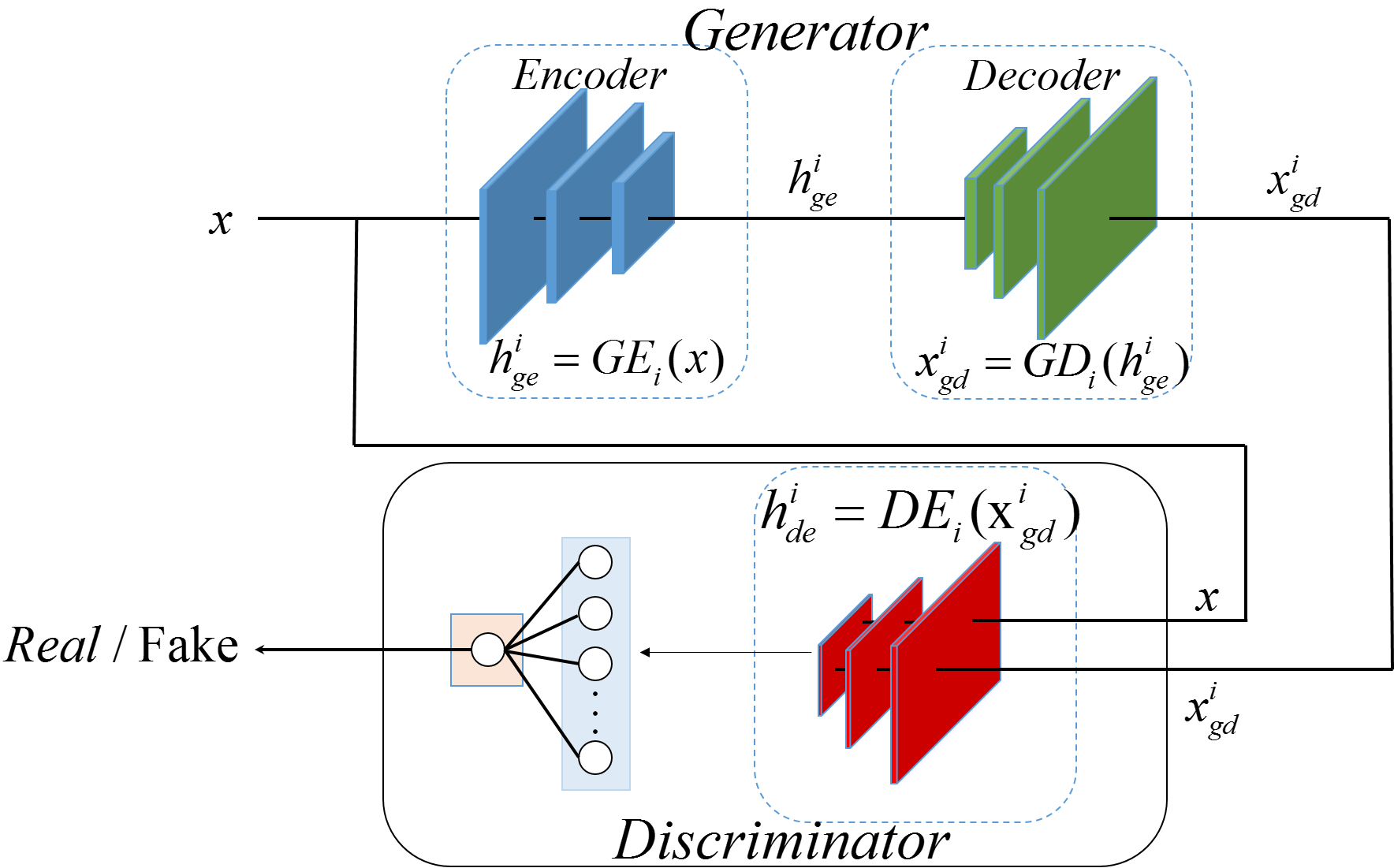}
\caption{The structure of GANs}
\label{fig:gan_f}
\end{figure}

The proposed network applies two GANs in parallel, one of which is responsible for learning patterns of normal samples ($GAN_{normal}$) and the other is for icing samples ($GAN_{icing}$). Two GANs share the same structure as shown in Fig \ref{fig:gan_f}. They are expressed as follows:

\begin{table}[hbt]
 \begin{threeparttable}
  \renewcommand\arraystretch{1}
\caption{Network architecture of GAN}
 \setlength{\tabcolsep}{4.5mm}
\label{tb:GANpara}
\center
\begin{tabular}{ll}
\hline
Layer        & Output size (F/Ks/S)\tnote{1}   \\ \hline
\hline
$\bf{Generator Encoder}$  \\
Input       & 1x1x28    \\
Conv (LeakyReLU/Batchnorm) & 4x1x25(4/4/1) \\
Conv & 8x1x22(8/4/1) \\
$\bf{Generator Decoder}$  \\
ConvTran (LeakyReLU/Batchnorm)       & 8x1x25 (8/4/1)    \\
ConvTran (LeakyReLU/Batchnorm)       & 1x1x28 (1/4/1)    \\
Tanh  & 1x1x28\\
$\bf{Discriminator}$ \\
Input       & 1x1x28    \\
Conv (LeakyReLU/Batchnorm) & 4x1x25(4/4/1) \\
Conv & 8x1x22(8/4/1) \\
FC       & 1    \\
Sigmoid & 1   \\
\hline  
\end{tabular}
\label{sgan}
 \begin{tablenotes}
     \item[1]Filters/Kernel size/Stride
   \end{tablenotes}
\end{threeparttable}
\end{table}

\begin{equation}
\begin{aligned}
h_{ge}^i &= GE_i(x)\\
x_{gd}^i &= GD_i(h_{ge}^i)\\
h_{de}^i &= DE_i(x_{gd}^i)\\
y^i &= h_{de}^i - h_{ge}^i\\
\end{aligned}
\end{equation}
where $i\in \{n, ic\}$ denotes normal and icing cases, respectively. $GE(\cdot)$, $GD(\cdot)$ and $DE(\cdot)$ present encoders and decoders in the generator and discriminator, respectively. $x_{gd}^i\in \mathcal{R}^{1\times28}$ is the synthetic sample. $h_{ge}^i\in \mathcal{R}^{8\times1\times22}$ and $h_{de}^i\in \mathcal{R}^{8\times1\times22}$ are extracted feartures of original and synthetic samples, respectively.  $y^i\in \mathcal{R}^{8\times1\times22}$ are outputs of GANs. The detailed structure of GANs can be found in Table~\ref{sgan}.

\begin{figure}[ht!]
\centering
\includegraphics[scale=0.28]{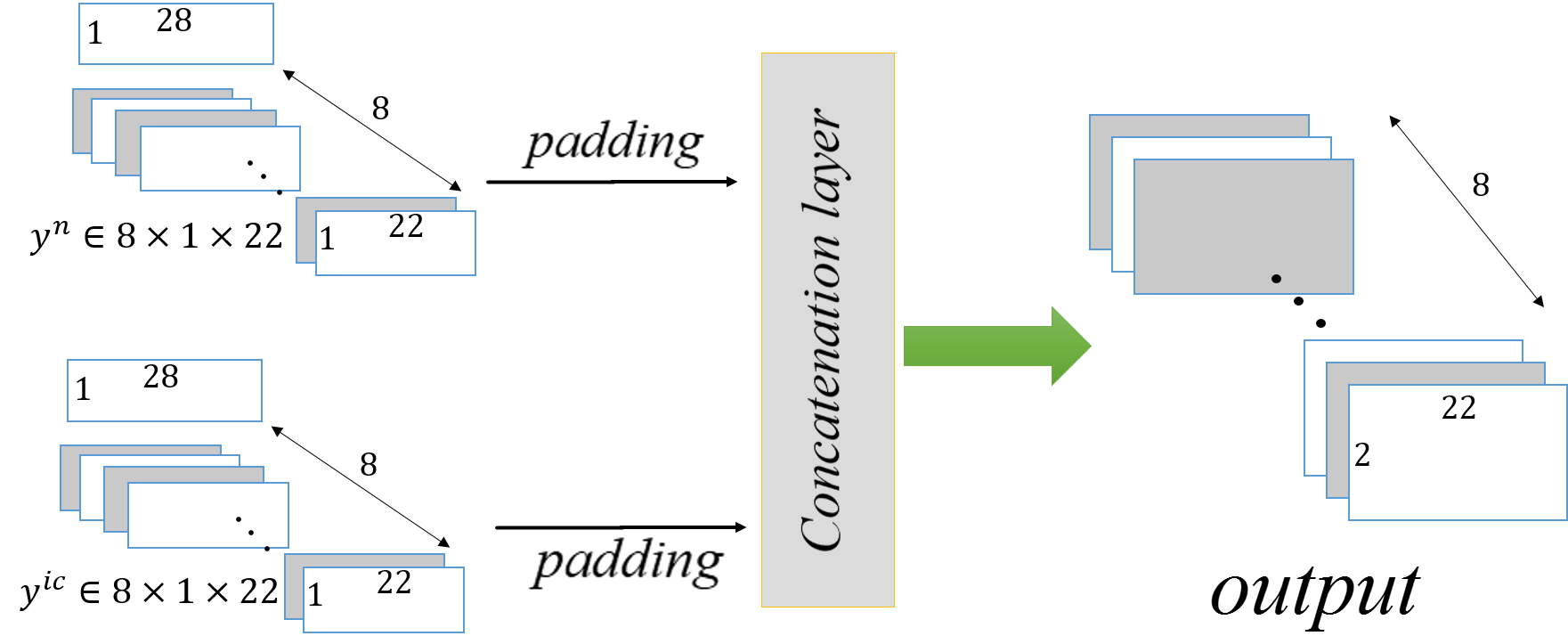}
\caption{The structure of concatenation layer}
\label{fig:conc}
\end{figure}

Output $y^i$ is a tensor composed of feature differences between original and synthetic samples. All outputs of two GANs are passed to the concatenation layer $CON(\cdot,\cdot)$ shown in Fig~\ref{fig:conc}. This layer combines $y^i$ to produce $8$ channels of $2\times22$ matrices $F$:
\begin{equation}
 F(i,:,:)=\left[y^n(i,:,:);y^{ic}(i,:,:)\right], 1\leq i\leq 8.
\end{equation}
where $i$ denotes the index for channels. The spacial information of concatenated outputs contains important patterns of inputs that are essential for classification. 

CNNs are well-known by its power to explore spatial information of samples~\cite{cnn11,cnn12,cnn13}. The characteristics of inputs are decomposed layer-by-layer through the network where each layer is focused on learning a particular type of features. To classify the states of wind turbine blade, concatenated signals $F$ are passed to the CNN classifier for anomaly detection:
\begin{equation}
Y = CNN(F)
\end{equation}
where $Y\in \mathcal{R}^{1\times2}$ is a vector whose elements present probabilities of normal and abnormal icing classes. The detailed structure of the CNN is introduced in Table~\ref{scnn}.

\begin{table}[hbt]
 \begin{threeparttable}
  \renewcommand\arraystretch{1}
\caption{Network architecture of CNN}
 \setlength{\tabcolsep}{3.5mm}
\label{tb:CNNpara}
\center
\begin{tabular}{ll}
\hline
Layer        & Output size (F/Ks/S)\tnote{1}   \\ \hline
\hline
Input       & 8x2x22    \\
Conv2d (LeakyReLU/Batchnorm) & 4x2x19(4/(1,4)/(1,1)) \\
FC       & 2   \\
Softmax  & 2    \\
\hline 
\end{tabular}
\label{scnn}
 \begin{tablenotes}
     \item[1]Filters/Kernel size/Stride
   \end{tablenotes}
\end{threeparttable}
\end{table}

To conclude, the proposed network in~\eqref{wn} can be further expanded as:
\begin{equation}
\begin{aligned}
Y = CNN\left\{CON\left[GAN_n(x),GAN_{ic}(x)\right]\right\}
\end{aligned}
\end{equation}

\subsubsection{Training procedure}

As the generator and the discrimiiator of a GAN are constructed with opposite purposes, they are updated in turn based on the game theory. In each iteration, one part remains unchanged whilst the other is updated to optimize current strategies against its opponent. The whole procedure advances until the Nash equilibrium is achieved. In the end, both parts learn their optimal strategies and reach a balance~\cite{goodfellow2014generative}.

Given a batch of samples, the optimization criteria for the generator and the discriminator are described as follows~\cite{jiang2019gan}:
\begin{equation}
\begin{aligned}
Generator\ G_i:\\
&\min L_{G_i}(x)\\
s.t.&\\
L_{G_i}(x) = \omega_{c}*&L_{con}(x) + \omega_{a}*L_{adv}(x) +\omega_{f}* L_{f}(x)\\
L_{con}(x) =& \frac{1}{N_i}\sum_{j=1}^{N_i}\|x^i_j-G_i(x^i_j)\|_2\\
L_{adv}(x) =& \frac{1}{N_i}\sum_{j=1}^{N_i}\log(1-D_i(G_i(x^i_j)))\\
L_f(x) =& \frac{1}{N_i}\sum_{j=1}^{N_i}\|GE_i(x^i_j)-DE_i(G_i(x^i_j))\|_2\\
Discriminator\ D_i:\\
&\min L_{D_i}(x)\\
s.t.&\\
L_D(x) = \frac{1}{N_i}\sum_{j=1}^{N_i}[&\log D_i(x^i_j)+\log(1-D_i(G_i(x^i_j)))]\\
\end{aligned}
\end{equation}
where $i\in \{n,ic\}$ denotes severity cases, $N_i$ is the batch size of class $i$ and $x^i_j\in \mathcal{R}^{1\times28}$ presents the $j$th sample of class $i$.

In the next step, training samples are propogated forward through the pre-trained GANs and the outputs are passed to the concatenation layer to generate feature tensors $F$ for classification.

As a classification problem, the cross-entropy loss is applied as the optimization criterion:
\begin{equation}
\begin{aligned}
&\min L_{CNN}(F)\\
s.t.&\\
L_{CNN}(F) =& \sum_i^{[n,ic]}\frac{1}{N_i}\sum_{j=1}^{N_i}-\log(CNN(F_j^i))
\end{aligned}
\label{tc}
\end{equation}
where $F^i_j$ represents the feature tensor of the $j$th sample in class $i$. Unlike GANs, samples of normal and icing classes are used to train the CNN at the same time. 

\subsection{PGANT: non-sufficient labeled data  scenario}

Enough high-quality labeled data is usually difficult to acquire, especially for industrial scenarios, not only because retrofitting enough sensors in packaged equipment is sometimes challenging, but also expensive and laborious labeling makes it daunting. For above reasons, transfer learning is increasingly becoming a promising tool to solve the basic problem of unlabeled and insufficient data under various operating conditions, through utilizing the model trained on data from source domain to improve learning performance of target domain. 

\subsubsection{Method overview}

\begin{figure}[ht!]
\centering
\includegraphics[scale=0.28]{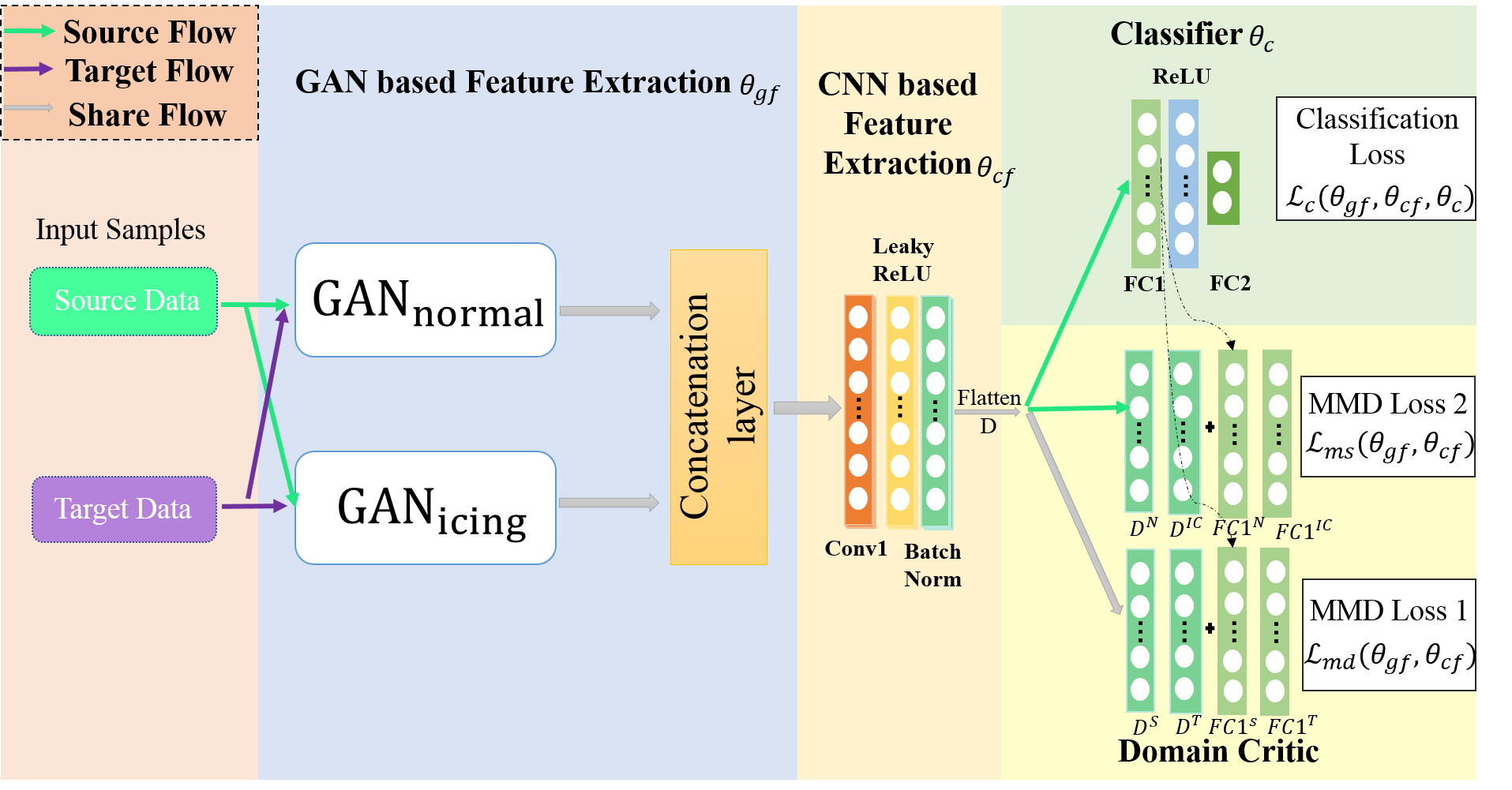}
\caption{The structure}
\label{fig:GTA}
\end{figure}

The neural network designed for icing detection of wind turbine blade without enough labeled data is composed of two GANs in paralell following by a classifier and a domain critic as shown in Fig \ref{fig:GTA}. The structures of two GANs and concatenation layer in this part are as the same as those in section \ref{tit:f1}. 

We define a source domain dataset $x^s={\left\{x^s_i\right\}}^{N^s}_{i=1}$ with labels $y^s={\left\{y^s_i\right\}}^{N^s}_{i=1}$, a target domain dataset $x^t = {\left\{x^t_i\right\}}^{N^t}_{i=1}$. Usually the feature space in source domain and target domain are the same ($x^s,x^t\in \chi$) but $P(x^s)\neq P(x^t)$, where $P(\cdot)$ represents a marginal probability distribution. In most cases, source domain samples are sufficient enough to learn an accurate classifier and with much larger data size than
the target domain, which means $N^s \gg N^t$.

There are four parts in our transfer learning framework, GAN based feature extractor, CNN based feature extractor, followed by classifier and domain critic. In the training stage, we first train $GAN_{normal}$ and $GAN_{icing}$ using normal and icing samples independently, and then initialize the two GANs in our proposed framework with the trained parameters. After concatenation layer, features $F$ extracted from GAN based neural networks are passed to CNN based feature extractor for further feature exploring: $D = cnn(F)$. As shown in Fig \ref{fig:GTA}, the CNN based feature extractor includes three layers: a convolution layer, a LeakyReLU activation layer, and a BatchNorm layer. For source domain data, a fully collected network is used to classify the feature vectors extracted through CNN based neural networks: $Y =FNN(D)$. In our proposed framework, the domain adaptation moddle includes a source-target domain distribution discrepancy and a normal-icing domain distribution discrepancy.

For a given input $x^s$ and $x^t$, the feed forward path can be summarized through the following equations,

\begin{alignat*}{2}
 y^n_s &= GAN_{normal}(x^s),&\quad  y^n_t &= GAN_{normal}(x^t) \\
 y^{ic}_s &= GAN_{icing}(x^s), &\quad  y^{ic}_t &= GAN_{icing}(x^t) \\
 F_s &= [y^n_s;y^{ic}_s],&\quad F_t &= [y^n_t;y^{ic}_t] \\
 D_s &= cnn(F_s),&\quad D_t &= cnn(F_t)\\ 
 Y_s &= FNN(D_s) 
\end{alignat*}

\subsubsection{Training procedure}
\label{sec:transfer optimization}
The proposed transfer learning framework has the following three optimization objects.
\begin{enumerate}[1)]
\item minimize the classification error in the source domain dataset;
\item minimize the MMD distance between the source and target domain dataset;
\item maximize the MMD distance between the normal and icing samples in source domain dataset.
\end{enumerate}

1)Object1: In order to accomplish transfer icing detection, our proposed framework should be able to distinguish icing cases from normal cases and learn domain-invariant features. Specifically, the classifier is designed to identify icing states. Therefore, the first optimization object is to minimize the classification error in the source domain dataset. To compute the difference between the predicted label $\widetilde{Y_i}$ and the ground truth ${Y_i}$ in the source domain, binary cross entropy loss function $\mathcal{L}_c$ is used to update the GAN based feature extractor, the CNN based feature extractor, and the classifier with parameters $\theta_{gf}$, $\theta_{cf}$ and $\theta_c$:
\begin{equation}
\label{eq:lc}
    \mathcal{L}_c = -\frac{1}{N^s}\sum_{i=1}^{N^s}{\sum_{j}^{|m,ic|}\mathbb{I}_{Y_i}(j)log(\widetilde{Y_i}(j))}
\end{equation}
where, $\mathbb{I}_{Y_i}$ is an indicator function for label ${Y_i}$. 

2) Object2: The domain critic module is designed to learn not only the domain-invariant features between the source domain and target domain but also state-distinguished vectors between normal and icing samples of the source domain. To solve the distribution difference between the source and target domain, we utilize MMD to learn invariant feature representation. Furthermore, two feature vectors are taken into consideration, features from CNN based feature extraction and features from the first fully connected layer, to improve the performance of MMD learning. Therefore, the distribution discrepancy distance between features learned from different domain can be written as:
\begin{equation}
\begin{aligned}
    \mathcal{L}_{md} = \bigg|\bigg|\frac{1}{N^s}\sum_{i=1}^{N^s}D_i^S-\frac{1}{N^T}\sum_{i=1}^{N^T}D_i^T\bigg|\bigg|_H^2\\
    +\bigg|\bigg|\frac{1}{N^S}\sum_{i=1}^{N^S}FC1_i^S-\frac{1}{N^T}\sum_{i=1}^{N^T}FC1_i^T\bigg|\bigg|_H^2\\
\end{aligned}
\end{equation}
where $N^T$ is the number of training samples from the target domain, and $||\cdot||_H$ is a reproducing kernel Hibert space. In practical, we use the unbiased estimation based on the Gaussian radial basis function (RBF), i.e., $k(x,y)=exp(||x-y||^2/2\sigma^2)$  \cite{tzeng2014deep} to complete the MMD computation.  

3) Object3: In our proposed network, state-distinguished vectors in source domain are learned not only by binary cross entropy loss in classifier, but also through MMD loss in domain critic. The optimization object in this part is to maximize the distribution discrepancy distance between normal and icing cases in source domain:
\begin{equation}
\begin{aligned}
    \mathcal{L}_{ms} = \bigg|\bigg|\frac{1}{N^N}\sum_{i=1}^{N^N}D_i^N-\frac{1}{N^{IC}}\sum_{i=1}^{N^{IC}}D_i^{IC}\bigg|\bigg|_H^2\\
    +\bigg|\bigg|\frac{1}{N^N}\sum_{i=1}^{N^N}FC1_i^N-\frac{1}{N^{IC}}\sum_{i=1}^{N^{IC}}FC1_i^{IC}\bigg|\bigg|_H^2\\
    \end{aligned}
\end{equation}
where $N^n$ is the number of normal samples from the source domain, and $N^{IC}$ is the number of icing samples from the source domain.

Combining those three optimization objects, the final objective function can be expressed in terms of the binary cross entropy loss $\mathcal{L}_c$ of the classifier according to Eq. \ref{eq:lc} and the distribution discrepancy distance $\mathcal{L}_{md}$ and $\mathcal{L}_{ms}$ associated with MMD of domain critic, i.e:
\begin{equation}
   \mathcal{L}(\theta_{gf}^*,\theta_{cf}^*, \theta_c^*) =\min_{\theta_{gf},\theta_{cf}, \theta_c }(\mathcal{L}_c-
    \alpha\mathcal{L}_{ms}+\beta\mathcal{L}_{md})
\end{equation}
where the hyperparameters $\alpha$ and $\beta$ determine how strong the
domain critic is. As shown in Fig \ref{fig:GTA}, $\theta_{gf}$, $\theta_{cf}$, and $\theta_c$ denote the parameters for GAN based feature extraction, CNN based feature extraction and classifier, respectively. 

We train the proposed method by Adam algorithm \cite{kingma2014adam},the parameters $\theta_{gf}$, $\theta_{cf}$, and $\theta_c$ are updated as follows:
\begin{equation}
\begin{aligned}
    \theta_{gf} &\leftarrow \theta_{gf}-\xi(\frac{\partial\mathcal{L}_c}{\partial\theta_{gf}}-\alpha\frac{\partial\mathcal{L}_{ms}}{\partial\theta_{gf}}+\beta\frac{\partial\mathcal{L}_{md}}{\partial\theta_{gf}})\\
    \theta_{cf} &\leftarrow \theta_{cf}-\xi(\frac{\partial\mathcal{L}_c}{\partial\theta_{cf}}-\alpha\frac{\partial\mathcal{L}_{ms}}{\partial\theta_{cf}}+\beta\frac{\partial\mathcal{L}_{md}}{\partial\theta_{cf}})\\
    \theta_{c} &\leftarrow \theta_{c}-\xi \frac{\partial \mathcal{L}_c}{\partial \theta_{c}}
\end{aligned}
\end{equation}
where $\xi$ is the learning rate.

\subsection{Two-Stage Training of Neural Networks}

Normally, neural networks are trained as an entirety using all training data. Practically, this training scheme may encounter two problems. With a large number of model parameters, it is extremely difficult to select good initial values to start with, which highly influences the performance of trained models. Since all model parameters are updated simutaneously, the algorithm requires many epochs to converge. In addition, given a large dataset, it takes long iterations for each epoch to finish a complete data scan.

This paper proposes a two-stage training scheme in order to improve model performance and reduce computational costs. To begin with, the training data are partitioned into two parts: normal and icing. In the first stage of training, components with differring functionality in the neural network are trained independently using different parts of the data. By updating fewer variables and using less data per time, model parameters are well initialized and the total computaional costs are effectively reduced. The second stage applies the pre-trained model in stage one as the basis and conducts a global optimization for the entire neural network so that the model performance can be further improved. 




\section{Experiments Results and Comparisons}
\subsection{Experiment Setup}

\subsubsection{Dataset Description}
Thanks to the first Industrial Big Data Innovation Competition (the Competition) in 2017, we have access to the real data from a wind farm in China. The raw data is collected from the supervisory control and data acquisition (SCADA) system which includes hundreds of dimensions every 7 seconds. According to the engineers' domain-specific knowledge, 26 continuous variables including real-time environmental data and wind turbine operating data relevant to the icing detection are preserved as the given multivariate signals. Detailed description of these variables is available online \cite{dataset_link}.

We choose three wind turbines' monitoring data for validating the performance of our proposed intelligent icing detection frameworks. For each wind turbine, we have icing and normal samples based on the fault records. Due to commercial confidentiality, all data given in the Competition have been encrypted, 
thus they lost the physical meaning of the original data in real conditions. Data details of three wind turbines are presented in Table \ref{tab:dd}.

We proposed two scenarios for GAN+CNN based framework and GAN+Transfer learning framework, respectively, they are (we split the dataset in Table \ref{tab:ed}):
\begin{enumerate}[1)]
\item 10\% of icing samples from Turbine A ($A\in[15,21,08]$) and the same number of normal samples, which are randomly chosen, as the training set. 40\% of icing samples from Turbine A and ten times than that from normal samples as the testing set; 
\item 60\% of icing samples from Turbine A and the same number of normal samples, which are randomly chosen with labels, as the source domain set. The same number of samples from Turbine B without labels as the target domain set. 40\% of icing samples of Turbine B and ten times than that of normal samples as the testing set. 
\end{enumerate}

\begin{table}[!ht]
\caption{Dataset Details}\label{tab:dd}
\begin{tabular*}{\hsize}{@{}@{}cccc@{}}
\toprule
Turbine No.& WT 15 &WT 21 &WT 8\\
\hline
Total Samples& 393886& 190494& 202328\\
\hline
Normal Samples& 350255(88.9\%)&  168930(88.7\%) & 180596(89.3\%)  \\
\hline
Icing samples& 23892(6.1\%)& 10638(5.6\%)& 11308(5.5\%)  \\
\hline
Invalid Samples& 19739(5.0\%)& 10926(5.7\%)& 10424(5.2\%) \\
\hline
Period & 2015.11 and 2015.12 &    2015.11 & 2015.11\\
\hline
\end{tabular*}
\end{table}

\begin{table}[!ht]
\caption{Experiments Design}\label{tab:ed}
\begin{tabular*}{\hsize}{ccc}
\toprule
Experiments &Training dataset &Testing dataset\\
\hline
 WT 15& Labeled 15: 10\%  &  WT 15: 40\%\\
\hline
 WT 21&  Labeled 21: 10\% &   WT 21: 40\%  \\
\hline
 WT 8&  Labeled 08: 10\% &  WT 8: 40\%  \\
\hline
 WT 15 - WT 21& Labeled 15: 60\% Unlabeled 21: 60\%& WT 21: 40\%\\
\hline
 WT 15 - WT 8 & Labeled 15: 60\% Unlabeled 8: 60\% &  WT 8: 40\%\\
\hline
 WT 21 - WT 8 & Labeled  21: 60\% Unlabeled 8: 60\% &  WT 8: 40\%\\
\hline
\end{tabular*}
\end{table}

\subsubsection{Data Preprocessing}

Data quality is often of vital importance to the performance of the constructed model. A uniform data format after data processing makes subsequent model training more conveniently manipulated. In our paper, we preprocess the data as follows: invalid data elimination, data balancing, new primary features extraction, and data normalization. 

About 5\% of the data in each wind turbine is invalid as shown in Table \ref{tab:dd}. They are unable to provide useful information and will have an adverse effect on the classification performance, so we directly eliminate them from the whole dataset. Besides, the number of normal samples is much more than that of icing samples, which makes our problem a typical case of imbalanced scenarios. When normal and icing labels are imbalanced, classifiers based on neural networks will ensure the accuracy of the majority classes by sacrificing the minority classes \cite{yin2015empirical}, which means the diagnostic results will bias towards normality for when testing. However, for icing detection in wind farms, our focus should be specifically on those minority icing classes. In this study, we adopt strong rule filter and down sampling to imbalance the dataset. The strong rule filter is that we only consider records where the wind power output is less than 2. Under the icing conditions, icing accretion on the wind turbines would cause changes of the blades surface airfoil, thus inevitably decreasing power coefficient and power output. Specifically, when a wind turbine blade is icing, it can not output relatively high power, which is consistent with the information we get from the three wind turbines: icing samples are almost impossible to appear under the condition when wind power output is more than 2. 

After balancing our dataset, a primary feature engineering is applied to help describe the  behavior of wind turbines and their operating environments. We first analyze the 26 variables selected by engineers in pairs. It can be found that because of the symmetrical mechanical structures of wind turbines, the pitch angle, pitch speed and motor temperature of three blades in one wind turbine at the same time are almost linear correlation. Therefore, we only consider the mean value of these three variables. We also notice that three ratios will show a growing trend during the icing period, which are added in the input variables:
\begin{equation}
\begin{aligned}
   \kappa_{w2p}&=(\frac{v_{ws} +5}{p+5})^2-1\\
    \kappa_{w2g}&=(\frac{v_{ws} +5}{v_{gs}+5})^2-1\\
    \kappa_{w2pg}&=(\kappa_{w2p}+1)(\kappa_{w2gs}+1)-1 
\end{aligned}
\end{equation}
where $v_{ws}, p, v_{gs}$ represent wind speed, power, and generator speed, respectively.

According to the model of Ice accretion on structures proposed by Makkonen \cite{makkonen2000models}, wind speed and temperature are two important factors contributing to icing blades, of which the most direct effect is that the output power is relatively low at the same wind speed. Therefore, significant factors related to temperature and power should also be considered, which are as follows: 
\begin{equation}
    \begin{aligned}
      \kappa_1 &= T_{int}-T_{env}, \quad \kappa_2 = \frac{p}{v_{gs}}, \quad \kappa_3 = \frac{p}{(v_{ws})^3}\\
      \kappa_4 &= \frac{\kappa_2}{v_{ws}^2}, \quad \kappa_5 = \frac{v_{gs}}{v_{ws}}
    \end{aligned}
\end{equation}

Thereafter, data normalization is conducted to eliminate the large magnitude difference between different variables without changing their distributions and interdependence. This can also speed up the learning efficiency of subsequent models. In this study, the following formula was used.
\begin{equation}
    \tilde{x} = y_{min}+(y_{max}-y_{min})*\frac{x-x_{min}}{x_{max}-x_{min}}
\end{equation}

where x and $\tilde{x}$ correspond to data before and after normalization, respectively, $x_{max}$ and $x_{min}$ correspond to the maximum and minimum values of the raw data, and $y_{max}$ and $y_{min}$ are the maximum and minimum values of the target range for the raw data. The target value of normalization in our paper is selected as $[-1,1]$ owing to the value range of the activation function tanh is $[-1, 1]$.



\subsubsection{Evaluation Criteria}
Because there are much more normal samples than abnormal icing samples in our case, accuracy is not enough to measure the model performance. The overall performance will be compared by using the area under curve (AUC) of the receiver operating characteristic (ROC). In ROC, the true
positive rate (TPR) is as a function of the false positive rate (FPR). TPR and FPR are defined as follows,
\begin{equation}
\begin{aligned}
    TPR&=\frac{TP}{TP+FN}\\
    FPR&=\frac{FP}{FP+TN}
\end{aligned}
\end{equation}
where TP is the number of positive samples predicted to be positive. FN is the number of positive samples predicted to be negative. FP is the number of negative samples predicted
to be positive, and TN is the number of negative samples predicted to be negative. AUC (the range is $[0,1]$) is not affected by class distribution, which is suitable for data sets with imbalanced class distribution. 

We also notice that engineers and maintainers are more concerned about the icing cases in real applications. That means they are more tolerant of false positives compared with false negatives. Therefore, we also consider a new score (utilized in the Competition) to help evaluate our proposed models:
\begin{equation}
    S = 1- \alpha*\frac{FN}{N_{normal}}-(1-\alpha)*\frac{FP}{N_{fault}}
\end{equation}
where $\alpha=\frac{N_{fault}}{N_{normal}}$. The range of S is $[0,1]$. The value is more close to 1, the better performance the model is.

\subsection{Experimental Analysis}
\subsubsection{PGANC Icing Diagnosis}

For comparing the performance of our proposed PGANC diagnosis framework, we consider the other two methods:

\begin{enumerate}[1)]
    \item \textbf{SVC}: Traditional machine learning method SVC is used for comparison;
    \item \textbf{CNN-based network}: To present ablation analysis of two GAN-based networks in our proposed network, we substitute them with a CNN-based network, and applied it on the data analysis of three wind turbines. The average result of 5 random initialization was taken as the final result. 
\end{enumerate}

Results corresponding to this experiment is shown in Table \ref{tab:com2}. From this table, we can see that our proposed two-stage PGANC framework outperforms all other methods. GAN-based networks get better performance means that GAN is better to capture intrinsic features in different classes (normal and icing cases in our paper) than CNN. One phenomenon noticed here is that stage 2 whole training performs better than stage 1 step training.  

\begin{table}[!ht]
\centering
\fontsize{8.5}{10}\selectfont
\begin{threeparttable}
\caption{Comparison}\label{tab:com2}
\begin{tabular}{cccccccccc}
    \toprule
    \multirow{2}{*}{Method}&
    \multicolumn{2}{c}{15}&\multicolumn{2}{c}{21}&\multicolumn{2}{c}{8}\cr
    \cmidrule(lr){2-3} \cmidrule(lr){4-5} \cmidrule(lr){6-7}
    &Score &AUC &Score &AUC &Score &AUC \cr
    \midrule
    SVC& 0.840 & 0.888 &0.915  &0.931 & 0.860&0.897\cr
    CNN& 0.839& 0.882  & 0.904  &0.919&0.869&0.880\cr
    PGANC stage1& 0.776& 0.864& 0.836 &0.883 &0.830 &0.848\cr
    PGANC stage2& \textbf{0.884}&\textbf{0.908} &\textbf{0.926}&\textbf{0.940}&\textbf{0.915}  &\textbf{0.933}\cr
    \bottomrule
    \end{tabular}
    \end{threeparttable}
\end{table}

\begin{table}[!ht]
\centering
\fontsize{8.5}{10}\selectfont
\begin{threeparttable}
\caption{Comparison2, training 5\%, testing 5:1}\label{tab:com22}
\begin{tabular}{cccccccccc}
    \toprule
    \multirow{2}{*}{Method}&
    \multicolumn{2}{c}{15}&\multicolumn{2}{c}{21}&\multicolumn{2}{c}{8}\cr
    \cmidrule(lr){2-3} \cmidrule(lr){4-5} \cmidrule(lr){6-7}
    &Score &AUC &Score &AUC &Score &AUC \cr
    \midrule
    SVC& 0.859 & 0.884 &0.925  &0.944 & 0.903&0.923\cr
    CNN& 0.863& 0.884  & 0.900  &0.927&0.909&0.926\cr
    PGANC stage1& 0.811& 0.851& 0.871 &0.904 &0.851 &0.878\cr
    PGANC stage2& \textbf{0.876}&\textbf{0.896} &\textbf{0.924}&\textbf{0.943}&\textbf{0.945}  &\textbf{0.949}\cr
    \bottomrule
    \end{tabular}
    \end{threeparttable}
\end{table}


\subsubsection{PGANT Icing Diagnosis}
For comparing the performance of our proposed PGANT diagnosis framework, we consider KNN, SVC, XGBOOST, PGANC diagnosis network. In this experiment, above methods are trained using labeled source domain data, and tested on target domain data.

Results corresponding to this experiment is shown in Table \ref{tab:com1}. From this table, we can see that the performance of our proposed network is  state-of-the-art in all comparing methods in terms of score evaluation. Although MCC of SVC and XGBOOST is sometimes show better performance, our proposed network is able to show comparing results. We should notice that PGANC network is also trained on source domain data, but it still gets better transfer performance on target domain data, which means that the GAN-based feature extraction not only captures our interesting characteristics in icing samples, but also show better generalization performance with higher robust ability given data from new turbines.

\begin{table}[!ht]
\centering
\fontsize{8.5}{10}\selectfont
\begin{threeparttable}
\caption{Comparison}\label{tab:com1}
\begin{tabular}{cccccccccc}
    \toprule
    \multirow{2}{*}{Method}&
    \multicolumn{2}{c}{15-21}&\multicolumn{2}{c}{15-8} &\multicolumn{2}{c}{21-8}\cr
    \cmidrule(lr){2-3} \cmidrule(lr){4-5} \cmidrule(lr){6-7}
    &Score  &AUC &Score  &AUC &Score  &AUC\cr
    \midrule
    KNN& 0.401& 0.659&0.341  &0.595 &0.487 &0.656\cr
    SVC& 0.676 & 0.810&0.629 &0.739 &0.770 &0.810\cr
    XGBOOST& 0.481& 0.708  &0.325&0.590 &0.538 &0.742\cr
    PGANC& 0.715& 0.823& 0.654&0.723&0.805 &\textbf{0.815}\cr
    PGANT & \textbf{0.769}& \textbf{0.849} &\textbf{0.779} &\textbf{0.779} &\textbf{0.819}& 0.800\cr
    \bottomrule
    \end{tabular}
    \end{threeparttable}
\end{table}

For comparing the performance of our proposed PGANT diagnosis framework, we also consider the ablation analysis of one addition MMD loss and GAN-based networks. In this part, we add Matthews Correlation Coefficient (MCC) \cite{alexandridis2014music} as one of our evaluation criteria, which can show whether the minority class could be predicted correctly from another perspective. The experiments are set as follows: 


\begin{enumerate}[1)]
    \item \textbf{GAN CNN+2 LOSS}: Method proposed in this paper; 
    \item \textbf{GAN CNN+1 LOSS}: In this method, all the networks are trained as described in Sec. \ref{sec:transfer optimization}, except trained without $\mathcal{L}_{ms}$;
    \item \textbf{CNN+2 LOSS}: In this method, GAN-based feature extraction is substituted by CNN-based networks and trained with two MMD loss $\mathcal{L}_{ms}$ and $\mathcal{L}_{md}$.
    \item \textbf{CNN+1 LOSS}: This framework is the same as \textbf{CNN+2 LOSS}, but trained without $\mathcal{L}_{ms}$. 
\end{enumerate}

\begin{figure}[ht!]
\centering
\includegraphics[scale=0.5]{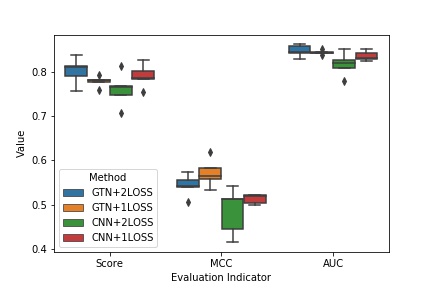}
\caption{Comparisons for the ablation study}
\label{fig:mc}
\end{figure}

We report the Evaluation Criteria averaged over five randomized trials. Results corresponding to this experiment is shown in Fig \ref{fig:mc}. From this figure, we can see that score and AUC outperform other methods, but for MCC PGANC with one MMD loss gets better performance.  

\section{Conclusion}
To achieve intelligent icing diagnosis of wind turbine, we proposed PGANC and PGANT for sufficient and non-sufficient labeled data of target turbines, respectively. Two parallel GANs are used for distinguished feature capture of icing and normal samples, and two-stage training process is adopted to further improve training efficiency and performance. Our proposed methods are tested on real SCADA data from a wind farm. Because icing cases of wind turbines are much fewer than normal cases, we take MCC, AUC, and a score which focuses more on the false negative rate than the false positive rate as evaluation criteria. Empirical results show that 1) PGANC outperforms traditional CNNs, and PGANT achieves higher robustness for icing diagnosis for different wind turbines. 2) our proposed two-stage training contributes to higher accuracy and efficiency.

\bibliographystyle{plain}
\bibliography{re}

\begin{thebibliography}{10}

\bibitem{IEC}
British standards institution. overhead transmission lines-design criteria. iec
  60826: 2017.
\newblock Available at: https://webstore.ansi.org/Standards/BSI/BSIEC608262017
  (Accessed on 19 December 2020).

\bibitem{dataset_link}
Detailed description for scada variables of wind turbines.
\newblock Available at: http://www.industrial-bigdata.com/Title, (Accessed
  December 10, 2020).

\bibitem{ganzoo}
The gan zoo: A list of all named gans.
\newblock Available at: https://github.com/hindupuravinash/the-gan-zoo
  (Accessed: 13th December, 2020).

\bibitem{GWEC}
Global wind energy council (gwec).
\newblock Available online: http://gwec.net/global-figures/graphs/ (accessed on
  20 June 2018).

\bibitem{alexandridis2014music}
Alex Alexandridis, Eva Chondrodima, Georgia Paivana, Marios Stogiannos, Elias
  Zois, and Haralambos Sarimveis.
\newblock Music genre classification using radial basis function networks and
  particle swarm optimization.
\newblock In {\em 2014 6th Computer Science and Electronic Engineering
  Conference (CEEC)}, pages 35--40. IEEE, 2014.

\bibitem{battisti2015wind}
Lorenzo Battisti.
\newblock {\em Wind turbines in cold climates: Icing impacts and mitigation
  systems}.
\newblock Springer, 2015.

\bibitem{cao2016study}
Mengnan Cao, Yingning Qiu, Yanhui Feng, Hao Wang, and Dan Li.
\newblock Study of wind turbine fault diagnosis based on unscented kalman
  filter and scada data.
\newblock {\em Energies}, 9(10):847, 2016.

\bibitem{chen2020robust}
Siyuan Chen, Yuquan Meng, Haichuan Tang, Yin Tian, Niao He, and Chenhui Shao.
\newblock Robust deep learning-based diagnosis of mixed faults in rotating
  machinery.
\newblock {\em IEEE/ASME Transactions on Mechatronics}, 25(5):2167--2176, 2020.

\bibitem{cheng2019wasserstein}
Cheng Cheng, Beitong Zhou, Guijun Ma, Dongrui Wu, and Ye~Yuan.
\newblock Wasserstein distance based deep adversarial transfer learning for
  intelligent fault diagnosis.
\newblock {\em arXiv preprint arXiv:1903.06753}, 2019.

\bibitem{dai2018ageing}
Juchuan Dai, Wenxian Yang, Junwei Cao, Deshun Liu, and Xing Long.
\newblock Ageing assessment of a wind turbine over time by interpreting wind
  farm scada data.
\newblock {\em Renewable Energy}, 116:199--208, 2018.

\bibitem{dao2018condition}
Phong~B Dao, Wieslaw~J Staszewski, Tomasz Barszcz, and Tadeusz Uhl.
\newblock Condition monitoring and fault detection in wind turbines based on
  cointegration analysis of scada data.
\newblock {\em Renewable Energy}, 116:107--122, 2018.

\bibitem{davis2016ice}
Neil~N Davis, {\O}yvind Byrkjedal, Andrea~N Hahmann, Niels-Erik Clausen, and
  Mark {\v{Z}}agar.
\newblock Ice detection on wind turbines using the observed power curve.
\newblock {\em Wind Energy}, 19(6):999--1010, 2016.

\bibitem{dong2020blades}
Xinghui Dong, Di~Gao, Jia Li, Zhang Jincao, and Kai Zheng.
\newblock Blades icing identification model of wind turbines based on scada
  data.
\newblock {\em Renewable Energy}, 162:575--586, 2020.

\bibitem{gantasala2016influence}
Sudhakar Gantasala, Jean-Claude Luneno, and Jan-Olov Aidanp{\"a}{\"a}.
\newblock Influence of icing on the modal behavior of wind turbine blades.
\newblock {\em Energies}, 9(11):862, 2016.

\bibitem{gao2020semi}
Yiping Gao, Liang Gao, Xinyu Li, and Xuguo Yan.
\newblock A semi-supervised convolutional neural network-based method for steel
  surface defect recognition.
\newblock {\em Robotics and Computer-Integrated Manufacturing}, 61:101825,
  2020.

\bibitem{ge2017prediction}
Yangming Ge, Dong Yue, and Lei Chen.
\newblock Prediction of wind turbine blades icing based on mbk-smote and random
  forest in imbalanced data set.
\newblock In {\em 2017 IEEE Conference on Energy Internet and Energy System
  Integration (EI2)}, pages 1--6. IEEE, 2017.

\bibitem{goodfellow2014generative}
Ian Goodfellow, Jean Pouget-Abadie, Mehdi Mirza, Bing Xu, David Warde-Farley,
  Sherjil Ozair, Aaron Courville, and Yoshua Bengio.
\newblock Generative adversarial nets.
\newblock In {\em Advances in Neural Information Processing Systems}, pages
  2672--2680, 2014.

\bibitem{gretton2012kernel}
Arthur Gretton, Karsten~M Borgwardt, Malte~J Rasch, Bernhard Sch{\"o}lkopf, and
  Alexander Smola.
\newblock A kernel two-sample test.
\newblock {\em The Journal of Machine Learning Research}, 13(1):723--773, 2012.

\bibitem{guangfei2018ice}
Zhou Guangfei, Tan Wen, and Zhang Da.
\newblock Ice detection for wind turbine blades based on pso-svm method.
\newblock In {\em Journal of Physics: Conference Series}, volume 1087, page
  022036. IOP Publishing, 2018.

\bibitem{habibi2019reliability}
Hamed Habibi, Ian Howard, and Silvio Simani.
\newblock Reliability improvement of wind turbine power generation using
  model-based fault detection and fault tolerant control: A review.
\newblock {\em Renewable Energy}, 135:877--896, 2019.

\bibitem{han2012scaled}
Yiqiang Han, Jose Palacios, and Sven Schmitz.
\newblock Scaled ice accretion experiments on a rotating wind turbine blade.
\newblock {\em Journal of Wind Engineering and Industrial Aerodynamics},
  109:55--67, 2012.

\bibitem{homola2006ice}
Matthew~C Homola, Per~J Nicklasson, and Per~A Sundsb{\o}.
\newblock Ice sensors for wind turbines.
\newblock {\em Cold regions science and technology}, 46(2):125--131, 2006.

\bibitem{jiang2019gan}
Wenqian Jiang, Yang Hong, Beitong Zhou, Xin He, and Cheng Cheng.
\newblock A gan-based anomaly detection approach for imbalanced industrial time
  series.
\newblock {\em IEEE Access}, 7:143608--143619, 2019.

\bibitem{khan2019survey}
Asifullah Khan, Anabia Sohail, Umme Zahoora, and Aqsa~Saeed Qureshi.
\newblock A survey of the recent architectures of deep convolutional neural
  networks.
\newblock {\em arXiv preprint arXiv:1901.06032}, 2019.

\bibitem{kingma2014adam}
Diederik~P Kingma and Jimmy Ba.
\newblock Adam: A method for stochastic optimization.
\newblock {\em arXiv preprint arXiv:1412.6980}, 2014.

\bibitem{lecun1989handwritten}
Yann LeCun, Bernhard Boser, John Denker, Donnie Henderson, R~Howard, Wayne
  Hubbard, and Lawrence Jackel.
\newblock Handwritten digit recognition with a back-propagation network.
\newblock {\em Advances in neural information processing systems}, 2:396--404,
  1989.

\bibitem{lecun1989backpropagation}
Yann LeCun, Bernhard Boser, John~S Denker, Donnie Henderson, Richard~E Howard,
  Wayne Hubbard, and Lawrence~D Jackel.
\newblock Backpropagation applied to handwritten zip code recognition.
\newblock {\em Neural computation}, 1(4):541--551, 1989.

\bibitem{liu2016dislocated}
Ruonan Liu, Guotao Meng, Boyuan Yang, Chuang Sun, and Xuefeng Chen.
\newblock Dislocated time series convolutional neural architecture: An
  intelligent fault diagnosis approach for electric machine.
\newblock {\em IEEE Transactions on Industrial Informatics}, 13(3):1310--1320,
  2016.

\bibitem{long2015learning}
Mingsheng Long, Yue Cao, Jianmin Wang, and Michael Jordan.
\newblock Learning transferable features with deep adaptation networks.
\newblock In {\em International conference on machine learning}, pages 97--105.
  PMLR, 2015.

\bibitem{makkonen2000models}
Lasse Makkonen.
\newblock Models for the growth of rime, glaze, icicles and wet snow on
  structures.
\newblock {\em Philosophical Transactions of the Royal Society of London.
  Series A: Mathematical, Physical and Engineering Sciences},
  358(1776):2913--2939, 2000.

\bibitem{ningbo2018ice}
LI~Ningbo, YAN Tao, LI~Naipeng, KONG Detong, LIU Qingchao, and LEI Yaguo.
\newblock Ice detection method by using scada data on wind turbine blades.
\newblock {\em Power Generation Technology}, 39(1):58--62, 2018.

\bibitem{cnn13}
N.~{Passalis} and A.~{Tefas}.
\newblock Training lightweight deep convolutional neural networks using
  bag-of-features pooling.
\newblock {\em IEEE Transactions on Neural Networks and Learning Systems},
  30(6):1705--1715, 2019.

\bibitem{rizk2020hyperspectral}
Patrick Rizk, Nawal Al~Saleh, Rafic Younes, Adrian Ilinca, and Jihan Khoder.
\newblock Hyperspectral imaging applied for the detection of wind turbine blade
  damage and icing.
\newblock {\em Remote Sensing Applications: Society and Environment},
  18:100291, 2020.

\bibitem{saxe2011random}
Andrew~M Saxe, Pang~Wei Koh, Zhenghao Chen, Maneesh Bhand, Bipin Suresh, and
  Andrew~Y Ng.
\newblock On random weights and unsupervised feature learning.
\newblock In {\em ICML}, volume~2, page~6, 2011.

\bibitem{sejdinovic2013equivalence}
Dino Sejdinovic, Bharath Sriperumbudur, Arthur Gretton, and Kenji Fukumizu.
\newblock Equivalence of distance-based and rkhs-based statistics in hypothesis
  testing.
\newblock {\em The Annals of Statistics}, pages 2263--2291, 2013.

\bibitem{cnn12}
G.~{Shomron} and U.~{Weiser}.
\newblock Spatial correlation and value prediction in convolutional neural
  networks.
\newblock {\em IEEE Computer Architecture Letters}, 18(1):10--13, 2019.

\bibitem{shu2016influences}
L~Shu, X~Ren, Q~Hu, X~Jiang, J~Liang, G~Qiu, and HT~Li.
\newblock Influences of environmental parameters on icing characteristics and
  output power of small wind turbine.
\newblock {\em Proc. CSEE}, 36(11):5873--5878, 2016.

\bibitem{shu2018study}
Lichun Shu, Hantao Li, Qin Hu, Xingliang Jiang, Gang Qiu, Ghyslaine McClure,
  and Hang Yang.
\newblock Study of ice accretion feature and power characteristics of wind
  turbines at natural icing environment.
\newblock {\em Cold Regions Science and Technology}, 147:45--54, 2018.

\bibitem{skrimpas2016detection}
Georgios~Alexandros Skrimpas, Karolina Kleani, Nenad Mijatovic,
  Christian~Walsted Sweeney, Bogi~Bech Jensen, and Joachim Holboell.
\newblock Detection of icing on wind turbine blades by means of vibration and
  power curve analysis.
\newblock {\em Wind Energy}, 19(10):1819--1832, 2016.

\bibitem{song2018wind}
Zhe Song, Zijun Zhang, Yu~Jiang, and Jin Zhu.
\newblock Wind turbine health state monitoring based on a bayesian data-driven
  approach.
\newblock {\em Renewable energy}, 125:172--181, 2018.

\bibitem{tautz2016using}
Jannis Tautz-Weinert and Simon~J Watson.
\newblock Using scada data for wind turbine condition monitoring--a review.
\newblock {\em IET Renewable Power Generation}, 11(4):382--394, 2016.

\bibitem{tzeng2014deep}
Eric Tzeng, Judy Hoffman, Ning Zhang, Kate Saenko, and Trevor Darrell.
\newblock Deep domain confusion: Maximizing for domain invariance.
\newblock {\em arXiv preprint arXiv:1412.3474}, 2014.

\bibitem{gan13}
J.~{Wang}, S.~{Li}, B.~{Han}, Z.~{An}, H.~{Bao}, and S.~{Ji}.
\newblock Generalization of deep neural networks for imbalanced fault
  classification of machinery using generative adversarial networks.
\newblock {\em IEEE Access}, 7:111168--111180, 2019.

\bibitem{cnn11}
R.~{Xin}, J.~{Zhang}, and Y.~{Shao}.
\newblock Complex network classification with convolutional neural network.
\newblock {\em Tsinghua Science and Technology}, 25(4):447--457, 2020.

\bibitem{fgan2}
W.~{Yang}, C.~{Hui}, Z.~{Chen}, J.~{Xue}, and Q.~{Liao}.
\newblock Fv-gan: Finger vein representation using generative adversarial
  networks.
\newblock {\em IEEE Transactions on Information Forensics and Security},
  14(9):2512--2524, 2019.

\bibitem{yin2015empirical}
Hua Yin and Keke Gai.
\newblock An empirical study on preprocessing high-dimensional class-imbalanced
  data for classification.
\newblock In {\em 2015 IEEE 17th International Conference on High Performance
  Computing and Communications, 2015 IEEE 7th International Symposium on
  Cyberspace Safety and Security, and 2015 IEEE 12th International Conference
  on Embedded Software and Systems}, pages 1314--1319. IEEE, 2015.

\bibitem{yu2018crack}
Mengyao Yu, Sheng Fu, Yinbo Gao, Hao Zheng, and Yonggang Xu.
\newblock Crack detection of fan blade based on natural frequencies.
\newblock {\em International Journal of Rotating Machinery}, 2018, 2018.

\bibitem{fgan}
Shiqi Yu, Rijun Liao, Weizhi An, Haifeng Chen, Edel~B. García, Yongzhen Huang,
  and Norman Poh.
\newblock Gaitganv2: Invariant gait feature extraction using generative
  adversarial networks.
\newblock {\em Pattern Recognition}, 87:179 -- 189, 2019.

\bibitem{yuan2019waveletfcnn}
Binhang Yuan, Chen Wang, Fei Jiang, Mingsheng Long, S~Yu Philip, and Yuan Liu.
\newblock Waveletfcnn: A deep time series classification model for wind turbine
  blade icing detection.
\newblock 2019.

\bibitem{zhang2018ice}
Lijun Zhang, Kai Liu, Yufeng Wang, and Zachary~Bosire Omariba.
\newblock Ice detection model of wind turbine blades based on random forest
  classifier.
\newblock {\em Energies}, 11(10):2548, 2018.

\bibitem{fgan1}
M.~{Zhang}, M.~{Gong}, Y.~{Mao}, J.~{Li}, and Y.~{Wu}.
\newblock Unsupervised feature extraction in hyperspectral images based on
  wasserstein generative adversarial network.
\newblock {\em IEEE Transactions on Geoscience and Remote Sensing},
  57(5):2669--2688, 2019.

\bibitem{zhang2012international}
Sufang Zhang.
\newblock International competitiveness of china's wind turbine manufacturing
  industry and implications for future development.
\newblock {\em Renewable and Sustainable Energy Reviews}, 16(6):3903--3909,
  2012.

\end{thebibliography}
\end{document}